\documentclass{midl} 

\usepackage{mwe} 

\title[Modeling and Reversing Brain Lesions Using Diffusion Models]{Modeling and Reversing Brain Lesions Using Diffusion Models}

\usepackage{algpseudocode}
\usepackage{wrapfig}
\usepackage{booktabs}
\usepackage{caption}
\usepackage{amsmath}
\usepackage{algorithm}
\usepackage{algpseudocode}
\usepackage{amsfonts}
\usepackage{xcolor}

 \midlauthor{\Name{Omar Zamzam} \Email{zamzam@usc.edu}\\
  \Name{Haleh Akrami} \Email{akrami@usc.edu}\\
  \Name{Anand Joshi} \Email{ajoshi@usc.edu}\\
  \Name{Richard Leahy} \Email{leahy@usc.edu}\\
  \addr Ming Hsieh Department of Electrical and Computer Engineering\\ University of Southern California\\ Los Angeles, CA 90089, USA}

\begin{document}

\maketitle

\begin{abstract}

Brain lesions are abnormalities or injuries in brain tissue that are often detectable using magnetic resonance imaging (MRI), which reveals structural changes in the affected areas. This broad definition of brain lesions includes areas of the brain that are irreversibly damaged, as well as areas of brain tissue that are deformed as a result of lesion growth or swelling. Despite the importance of differentiating between damaged and deformed tissue, existing lesion segmentation methods overlook this distinction, labeling both of them as a single anomaly. In this work, we introduce a diffusion model-based framework for analyzing and reversing the brain lesion process. Our pipeline first segments abnormal regions in the brain, then estimates and reverses tissue deformations by restoring displaced tissue to its original position, isolating the core lesion area representing the initial damage. Finally, we inpaint the core lesion area to arrive at an estimation of the pre-lesion healthy brain. This proposed framework reverses a forward lesion growth process model that is well-established in biomechanical studies that model brain lesions. Our results demonstrate improved accuracy in lesion segmentation, characterization, and brain labeling compared to traditional methods, offering a robust tool for clinical and research applications in brain lesion analysis. Since pre-lesion healthy versions of abnormal brains are not available in any public dataset for validation of the reverse process, we simulate a forward model to synthesize multiple lesioned brain images.
\end{abstract}


\section{Introduction} 

A brain lesion refers to an abnormality or injury within brain tissue, often detectable by imaging modalities such as magnetic resonance imaging (MRI), where it appears as alterations in the structure, function, or appearance of the brain. Lesions vary considerably in nature, and their definition encompasses any abnormalities that appear in brain tissue - this includes regions of irreversibly damaged tissue (such as necrotic zones) and areas of tissue deformation caused by mass effects from tumor growth, edema, or pressure from surrounding proliferating tissue. The extent of functional impairment correlates not only with the volume of the lesion but also with the degree of displacement or compression exerted on adjacent structures by these mass effects \cite{sorribes2019biomechanical} \cite{greenspan1972models}.
Edema, in particular, often manifests as swelling in the surrounding white matter, amplifying the overall volume expansion caused by the lesion, which may lead to further structural and functional damage. Necrotic tissue is indicative of irreversible damage, while adjacent deformed tissue can sometimes recover function, provided the compression or displacement is alleviated \cite{wasserman1996patient}. Understanding the distinction between regions of permanently damaged tissue and those affected by potentially reversible deformation is paramount for accurate surgical planning. This differentiation enables the preservation of viable but deformed tissue while ensuring the removal of necrotic or irreversibly damaged regions.  In addition, this understanding facilitates monitoring postoperative recovery and assessing the brain's ability to heal.

In this study, we propose a diffusion model-based framework for analyzing and reversing the brain lesion process. The proposed end-to-end framework is designed to first segment anomalous regions in the brain, then estimate tissue deformations within the anomalous regions. Estimating the tissue deformations allows for applying the inverse of these deformations to restore displaced tissue to its original position. Reversing deformed tissue back to the original position reduces the identified anomaly to only irreversible tissue damage. We inpaint the remaining damaged tissue to arrive at an estimation of the pre-lesion healthy brain shape. Our method offers several advantages: it segments brain lesions and distinguishes between irreversibly damaged regions and those deformed due to neighboring tissue growth or damage. It provides more robust and accurate labeling of lesioned brains compared to existing direct methods, aids in quantifying the impact of lesion growth on all brain regions of interest, and accurately estimates the brain’s pre-lesion structure.

This proposed framework of reversing brain lesions assumes a forward process for lesion appearance and growth. This forward process is based on a model that is well-established in biomechanical studies that model brain lesions \cite{sorribes2019biomechanical} \cite{greenspan1972models} \cite{mohamed2006deformable} \cite{wasserman1996patient} where a brain region is damaged and then grows to deform and exert pressure on surrounding regions.

To validate our proposed lesion reversal framework, we simulate the assumed forward model to allow for the synthesis of multiple lesioned brain images that serve as examples with known-groundtruth and that can be used to assess the accuracy of our proposed framework.

\section{Related Work}

The general problems of lesion segmentation, intracranial pressure analysis in lesioned brains, and lesioned brain labeling have been extensively explored in the literature \cite{kansal2000simulated} \cite{nalepa2019data} \cite{sorribes2019biomechanical} \cite{wyatt2022anoddpm}. Early optimization-based methods provided holistic approaches to these problems \cite{mohamed2006deformable}, yet more recent learning-based methods have often examined these aspects separately. Given their ability to handle large datasets and provide fast processing, learning-based methods can offer complementary insights. By integrating biological models with statistical guidance from large datasets, these methods help address some of the limitations seen in purely model-based approaches, enabling more scalable solutions.

Many developed methods tackle the separate problem of brain lesion segmentation. Those methods can be categorized into three categories: 1) Classical methods; where edge detection, thresholding, and template-based methods are used to segment lesions. 2) Machine Learning methods; which rely on feature extraction from images based on experts knowledge and training a machine learning model to detect the lesion features. 3) Deep Learning methods; where supervised learning is used to train deep neural networks to segment lesions using fully labeled datasets, or unsupervised learning can be used to denoise input images to get normal images at the output, and then thresholding the difference image comparing input and output. \cite{jyothi2023deep} provides a comprehensive review on state-of-the-art lesion segmentation methods. These methods mainly lack a clear way of understanding the lesions in the brain where restoration of functionality is unfeasible (damaged tissue) or feasible (deformed normal tissue) or understanding of the effects of the lesion on areas of the brain that surround the lesion. 

Lesioned brain labeling is a problem where the goal is to map information from a brain atlas to a lesioned brain image by learning a mapping between the atlas and the subject brain. This problem has been classically solved in the literature by directly registering lesioned brains to atlases (in the same way a healthy brain would be labeled) and mapping information from the atlas to the lesioned brain. These methods in deformable image registration often operate under the assumption of the existence of a diffeomorphism between the images being aligned \cite{zou2022review,wu2022nodeo,sotiras2013deformable,hill2001medical}, however, this assumption does not hold in medical scenarios such as pathological image to atlas registration or longitudinal studies comparing pre-operative and post-operative images \cite{zacharaki2009non}. The early work in \cite{mohamed2006deformable} provides an approach for the deformable registration of brain atlases to lesioned brains, which is a valuable tool for analyzing lesioned brains. Their method relies on decomposing the deformation map between the brain atlas and lesioned brains into two orthogonal components: one capturing inter-subject anatomical variations and another representing tumor-induced mass effects. Using this decomposition, they parameterize the tumor's impact, estimate biomechanical model parameters, and simulate mass effects to construct deformation maps that enable atlas-based labeling.

While this approach for labeling lesioned brains has been influential and valuable, it faces certain limitations when applied to diverse or complex pathological scenarios. It relies on explicit parameterization of tumor-induced deformations, requiring assumptions about the lesion's location, size, and mechanical properties, which may not generalize well to diverse or atypical pathological scenarios. Additionally, the biomechanical model used in this method simplifies the complex and non-linear interactions between tumor growth and surrounding tissue, potentially leading to inaccuracies when the lesion exhibits irregular shapes or when substantial edema or other pathologies are present. The method's reliance on predefined model parameters also limits its scalability and adaptability to new datasets or imaging modalities, as recalibration is required for each new scenario.\\

In this work, we present a lesioned brain analysis pipeline, where a lesioned brain image is analyzed by segmenting lesions, estimating tissue deformation within lesions, and inpainting the image to arrive at an estimation of the pre-lesion healthy brain image. We show in section \ref{brain labeling} that our proposed lesion reversal framework provides enough information that can be used to label lesioned brains and provide better labeling quality compared to classical approaches.

We also present a simulation of the assumed forward model that models lesion appearance in brains. In the forward process, we allow for a precise specification of the location and size of the lesion in the brain, extent of lesion growth, and deformation of surrounding tissue.

The following sections of the paper are organized as follows: In section 3, we introduce the assumed lesion process forward model, then we introduce the proposed lesion reversal framework. In section 4, we cover the experiments and evaluations used to assess each component of the proposed methods. In section 5, we summarize the contributions presented in this work.

\section{Methods}

As the core contribution of this work is a framework for reversing the lesion process, it is essential to first establish a well-defined forward model of lesion development. This provides a principled foundation for formulating and validating the reversal process. Hence, we begin here by introducing the lesion process forward model that forms the basis of our proposed analysis framework.

\subsection{Lesion Process Simulation}\label{Lesion Synthesis Process}

The goal of this method is to simulate the lesion appearance process in the brain, to be able to synthesize lesioned brain images, for which the original healthy brain image is known. Synthesizing lesioned brain images involves a series of precise steps designed to generate realistic and varied lesion shapes. Initially, the lesion core is selected by determining a starting point and defining its size. Subsequently, the desired size for the final lesion area is specified. The lesion core is initialized within the brain image as a spherical, triangulated mesh, with intensity values averaged from the surrounding tissue to ensure realistic integration. Then, the mesh vertices are perturbed with Perlin noise, forming a more irregular, blob-like structure. This mesh was then transformed into a volumetric mask to represent the core lesion. A corresponding larger blob, representing the final lesion size, is then created. To model tissue displacement due to lesion growth, we co-registered the core lesion mask to the final lesion mask. The lesion growth was extrapolated to the rest of the brain by finding a dense displacement field (DDF) that models the displacement of each voxel under the constraint that the core lesion mask maps to the final lesion mask. To ensure that the DDF is diffeomorphic, we first found a dense velocity field (DVF) and integrated the DVF to find the DDF. The integration of DVF was performed by generating exponential and logarithmic maps of DVF to tangent space of diffeomorphisms. This approach is known to guarantee diffeomorphism and is known to have a simplified numerical solution (\cite{joshi2018rfdemons} and \cite{vercauteren2009diffeomorphic}). To model the tissue displacement, we used an incompressible fluid regularizer (divergence minimizing). Algorithm \ref{alg:lesion_synthesis} in Appendix section \ref{alg:lesion_synthesis} shows the steps of the lesion synthesis process.

\begin{figure}[tbh]
    \centering
\includegraphics[width=0.9\textwidth]{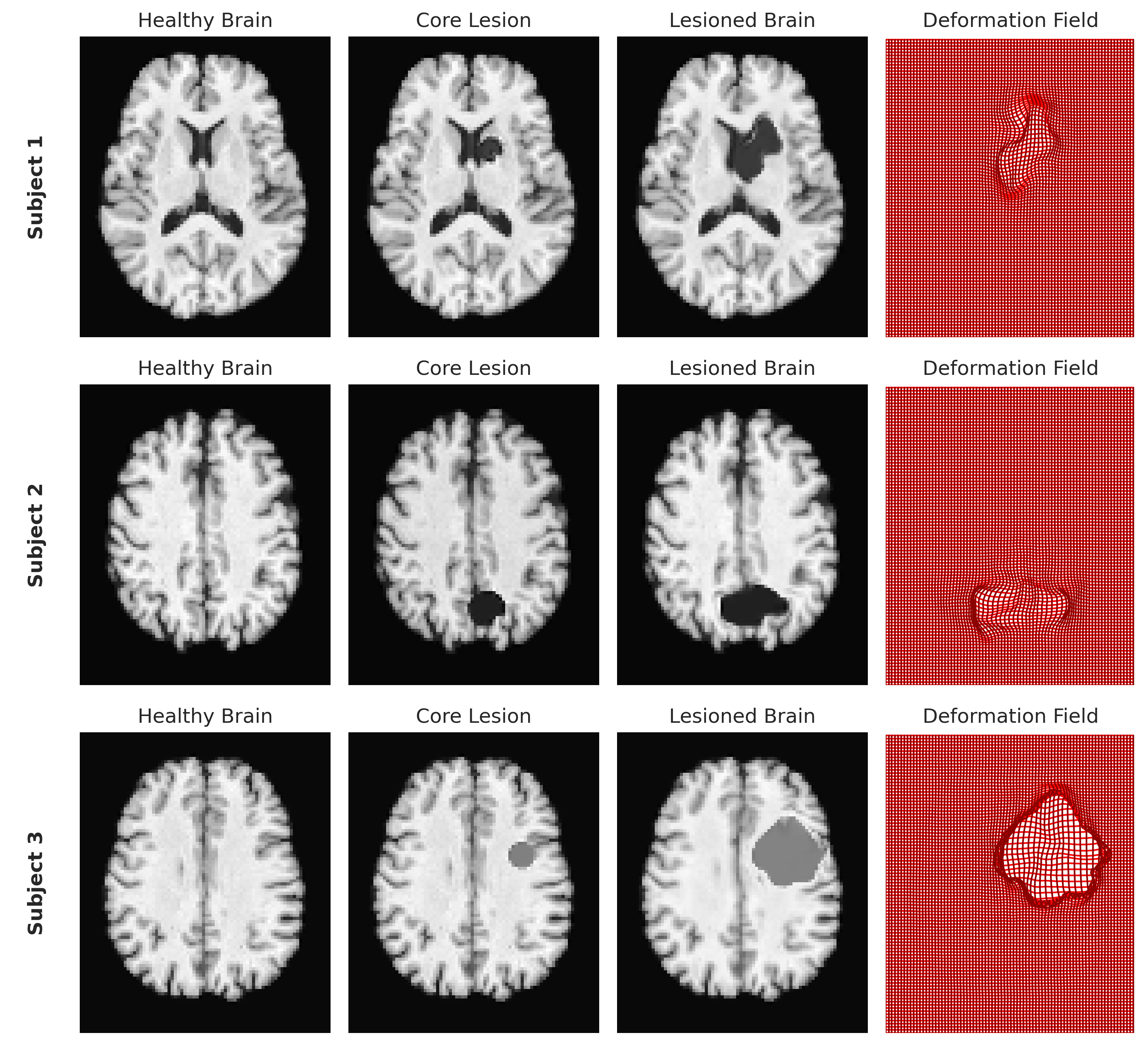}
    \caption{Lesion synthesis steps; adding a core lesion of a chosen size and adding it to a point in the white matter and then deforming the core lesion into a bigger lesion of a random shape using a nonlinear deformation process.}
    \label{fig:lesion_synthesis}
\end{figure}

The DVF in this process is generated by a neural network \(reg_{\phi}(\cdot)\), trained to align a moving brain image \(x\), which contains the core lesion, with a target brain image \(y\), which contains a larger lesion. The network outputs a DVF, which, when applied to the moving image, minimizes the mean squared error between the deformed image and the target image. Regularization based on the bending energy of the deformation field is incorporated to balance registration accuracy and the smoothness and physical plausibility of the generated deformations. The loss function used to train the network is

\begin{equation}
\mathcal{L}_{reg}(\phi) = \|y_{i}-reg_{\phi}(x_{i})\|_2^2 + \lambda \cdot \text{BendingEnergy}(reg)
\label{regEq}
\end{equation}
where \(\lambda\) is a parameter used to weight the regularization term. In this training, \(\lambda\) was chosen to be \(1\). This training, since it's done using only one brain image, it is done in seconds which allows for a speedy processing of the images. Figure \ref{fig:lesion_synthesis} shows example brain images on which the lesion synthesis process was applied. The method is implemented in Python, utilizing the MONAI and PyTorch libraries. The ADAM optimizer was used to minimize the loss function.

\subsection{Lesion Process Reversal}

Estimating the pre-lesion brain in a patient can be extremely valuable in monitoring post-surgical healing process, understand the effects the lesion had on surrounding tissue, and labeling the lesioned brain in fundamentally correct and tractable steps. We propose a pipeline for estimating the pre-lesion healthy brain by reversing the lesion process that happens by first having a damaged tissue or a tumor that starts with certain size and gradually grows to exert pressure on surrounding tissue, causing deformations. 

Starting from the lesioned brain (third column in Figure \ref{fig:lesion_synthesis}), we try to estimate the core lesion brain (second column in Figure \ref{fig:lesion_synthesis}) by estimating the inverse of the deformation field associated with the lesion growth (fourth column in Figure \ref{fig:lesion_synthesis}) and apply it to the lesioned brain image. After estimating the core lesion image, we use an inpainting diffusion model to replace the damaged tissue area in the brain by its normal counter part, finally arriving at the pre-lesion healthy brain image (first column in Figure \ref{fig:lesion_synthesis}).

Since the deformation process happening in the lesion process is local, i.e., it takes place only at the lesion area and the surrounding tissue and hardly has any effect on the rest of the brain, the first step of estimating it is the segmentation of the lesion area such that the estimated deformation field can be constrained to be local.

\begin{figure}[tbh]
    \centering
\includegraphics[width=\textwidth]{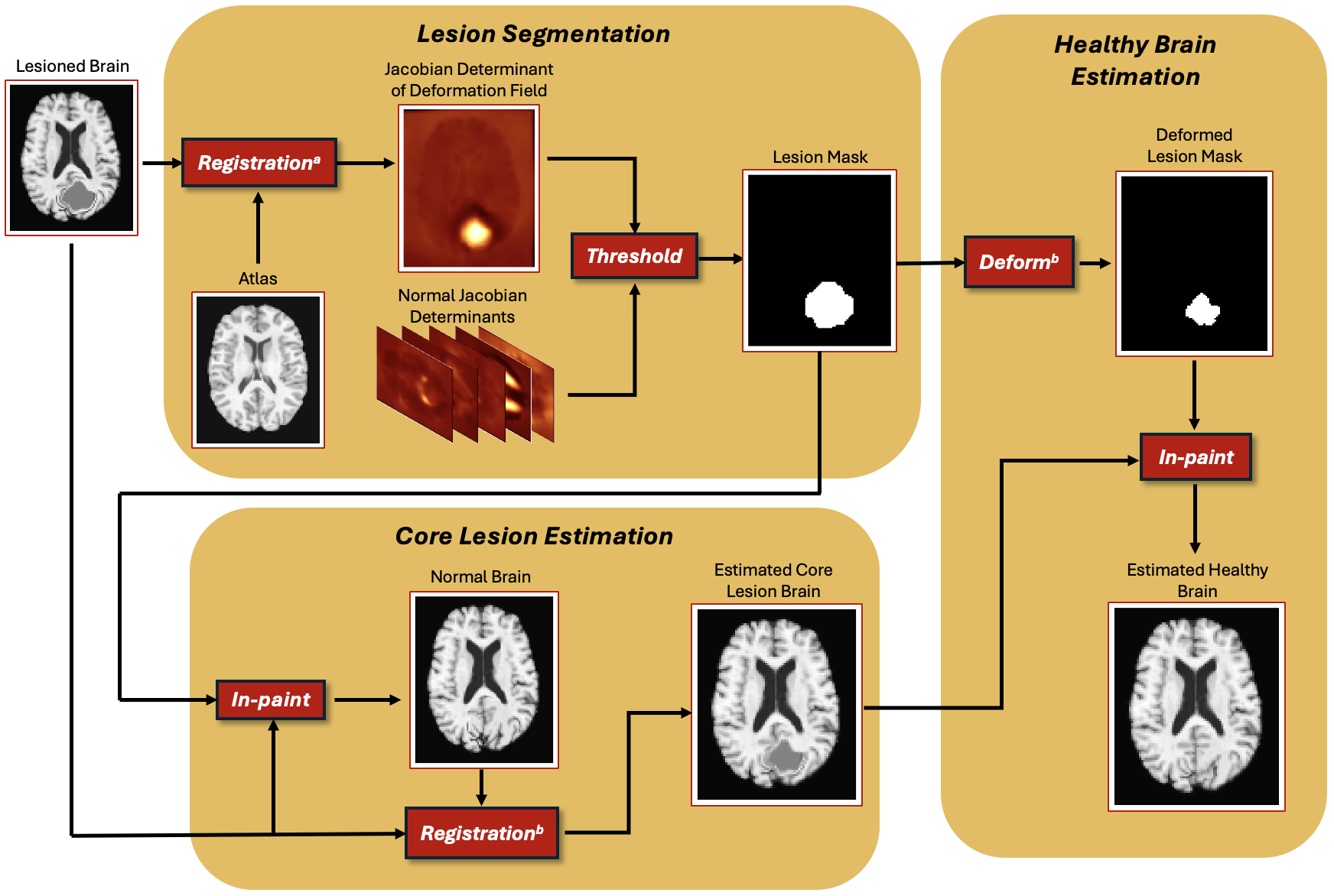}
    \caption{Lesion Process Reversal; Lesion segmentation (Top left); Registering the lesioned brain to an atlas, calculating the Jacobian determinant of the resulting deformation field, and then thresholding it against a pool of normative Jacobian determinants. Core lesion estimation (Bottom left); The lesioned brain is inpainted at the identified lesion mask, and then registered to the inpainted image, resulting in shrinkage of the lesion to its original shape before lesion grew. Healthy brain estimation (Right); The core lesion brain is inpainted at the deformed lesion mask.}
    \label{fig:Lesion_Reversal}
\end{figure}

\subsubsection{Lesion Segmentation}

Relying on the fact that the mappings between lesioned brain images and normal brains are particularly extreme around the lesion area due to severe structural abnormalities, we propose to identify the lesion area by registering the lesioned brain image to an atlas image. The most challenging regions to map to the atlas are then identified as potential lesion areas. Specifically, we detect extreme shrinkage or expansion in the deformation field that maps the atlas image to the lesioned brain image by thresholding the Jacobian determinant of the deformation field. This thresholding is performed using a validation set of \textit{normative} Jacobian determinants, ultimately producing a lesion segmentation mask.

The top left block in Figure \ref{fig:Lesion_Reversal} illustrates the steps involved in finding the lesion mask given the lesioned brain image.

The normative Jacobian determinant validation set is formed by training the registration network mentioned in \ref{regEq} to deform 100 T1-weighted normal brain images (from HCP dataset) to brain atlases, the network is then deployed on a validation set of 44 healthy subjects, where each brain is registered to an atlas. The pool of Jacobian determinants of each of the registration processes represents a validation set of normative Jacobian determinants used as a null distribution against which the registration of subjects at test time is compared for thresholding. The Jacobian determinant is used because it is an accurate measure of local shrinkage or expansion of the original volume during the deformation process. A growing tumor, for instance, will cause local deformation as it invades the space previously occupied by healthy brain tissue. This will be reflected in these deformation fields. Therefore, thresholding the Jacobian determinant is used to identify the lesion boundary (or boundaries if more than one) in the subject image after it has been deformed through registration to an atlas. The precision afforded by this approach ensures a robust basis for subsequent analyses in identifying and characterizing brain lesions.

\subsubsection{Estimating Core Lesion}

After estimating a lesion mask from the last step, the mask is used to estimate the \textit{local} deformation field that can be directly used to transform the lesion image to the core lesion image. We use an inpainting diffusion model to inpaint the lesioned brain image at the lesion mask to have a normal brain image that is identical to the lesioned brain everywhere except at the lesion area. Then, the lesioned brain is registered to the inpainted image. This registration results in a deformation field that does not change the lesioned brain anywhere except at the lesion part, where lesion cannot be deformed to any corresponding part in the normal brain image, and the deformed tissue surrounding the lesion can be deformed to its corresponding undeformed tissue.

The bottom left figure in Figure \ref{fig:Lesion_Reversal} shows an example of the steps followed in estimating the core lesion brain.

\subsubsection{Estimating pre-Lesion Brain}

Having the core lesion image from the last step along with the core lesion mask (found by applying the deformation field –that was applied to the lesioned brain– to the lesion mask), we use inpainting diffusion model to inpaint the core lesion with its healthy counterpart to arrive at the pre-lesion brain.

The right figure in Figure \ref{fig:Lesion_Reversal} shows an example of the steps followed in estimating the healthy brain given the core lesion brain and the lesion mask.

\subsection{In-painting Diffusion Model}

A core component of the lesion reversal process is the in-painting model. The in-painting model is essential in 1)  it replaces abnormalities identified by the lesion mask with normal tissue, enabling the deformed tissue to revert to its original position when the lesioned brain is registered to it, and in 2) It fills the core lesion mask with normal tissue, ensuring that the in-painted brain  represents a normal brain image that can be effectively processed using brain analysis software designed for normal brain structures.

We used a 3D diffusion model as an in-painting module to achieve a high quality in-painting and ensure the generation of consistent and realistic brain tissue, guided by the surrounding tissue.

We developed a guided in-painting module by fine-tuning the diffusion model for the inpainting task. Inspired by Stable-Diffusion-inpainting\footnote{https://huggingface.co/runwayml/stable-diffusion-inpainting}, we applied prompt tuning by adding additional input channels (for the encoding mask) to the U-Net whose weights were zero-initialized after restoring the non-inpainting checkpoint. For more coherent inpainting and using the non-masked information, we generated the noisy input as a combination of a noisy localized input and a non-masked input:

\begin{equation}
    \begin{aligned}
        \mathbf{x}_t &= \sqrt{\alpha_t} \mathbf{x}_{t-1} + \sqrt{1 - \alpha_t} \boldsymbol{\epsilon}, \\
        \mathbf{x^{'}}_t &= \mathbf{x}_t \odot m + \mathbf{x}_0 \odot (1-m)
    \end{aligned}
\end{equation}

\noindent
where $\mathbf{x^{'}}_t$ and the inpainting mask $m$ are the input to the network. 
We train a network $\boldsymbol{\epsilon}_\theta$ to predict the noise ${\epsilon}$ from the noisy $\mathbf{x}_t$:
\begin{equation}
    \mathcal{L}(\theta) = \mathbb{E}_{t, \mathbf{x}_0, \boldsymbol{\epsilon}} \left[ \| \boldsymbol{\epsilon} - \boldsymbol{\epsilon}_\theta(\mathbf{x^{'}}_t, t,m) \|^2 \right],
\end{equation}

We then combine the inpainting and the non-masked information in the reverse process. 

\begin{align}
    \mathbf{x}^{\text{unknown}}_{t-1} &= \frac{1}{\sqrt{\alpha_t}} 
    \left( 
        \mathbf{x}_t - \frac{1-\alpha_t}{\sqrt{1-\bar{\alpha}_t}} \boldsymbol{\epsilon}_\theta(\mathbf{x}_t, t) 
    \right), \\
    \mathbf{x}_{t-1} &= \mathbf{x}_0 \odot m + \mathbf{x}^{\text{unknown}}_{t-1} \odot (1 - m)
\end{align}

To allow the use of irregularly shaped masks, we developed a module for random 3D mask generation to use during training. The 3D mask is created by first forming a spherical mesh.  
To deform the mesh so that it has an irregular but natural appearance, Perlin noise is then applied to the position of each vertex. This results in a random, uneven surface. We then fill the sphere to create the 3D lesion-like image mask. 
We compared this module with a "RePaint" method \cite{lugmayr2022repaint} built on a pre-trained unconditional diffusion model.

\section{Experiments and Evaluations}

In this section, we present multiple experiments and evaluations that were implemented to: 1) Assess the accuracy of the deformation-based lesion segmentation method, and 2) Assess and compare the proposed in-painting diffusion model compared to SOTA in-painting models, and 3) Assess the usefulness of the lesion reversal process in enhancing the labeling of lesioned brains using the direct classical processing by brain software, and 4) Qualitatively analyze the ability of existing brain softwares to process real lesioned brains compared to normal brains.

\subsection{Datasets and Model Architectures}

For the in-painting diffusion model and registration model, we used a U-Net with three main levels. An attention mechanism is integrated into the third stage of the model. We trained the model using T1-weighted images from four publicly available datasets: Human Connectome Project (HCP) \footnote{https://humanconnectome.org}, Cambridge Centre for Ageing and Neuroscience (Cam-CAN) \footnote{https://www.cam-can.org}, the UK Biobank (UKB)\cite{sudlow2015uk}, along with the IXI \footnote{https://brain-development.org/ixi-dataset} dataset. Detailed information on all the datasets we used and preprocessing is provided in appendix \ref{preprocessing} and \ref{dataset}.

\subsection{Lesion Detection Evaluation}

We evaluate our lesion segmentation model performance on 100 images randomly selected from the BraTS21 T1W dataset (see appendix \ref{dataset} for dataset details). We compared the reconstruction errors obtained using our approach against three different diffusion models: 2D diffusion, 3D diffusion model, and latent 3D diffusion. For all three baseline models, we segmented the lesion by first adding noise and then denoising using the reverse process\cite{wyatt2022anoddpm}, and then thresholding the error image (difference between original and reconstructed images) based on the distribution of a healthy population validation set. We used the 95th percentile threshold computed from the healthy validation set error distributions for all baseline methods. For our deformation based method, we thresholded the Jacobian determinants using the 95th percentile as an upper limit and the 5th percentile as a lower limit. This is because both the high and low Jacobian determinants indicate an extreme deformation (expansion or contraction, respectively).

Since we are dealing with 3D images, the 2D diffusion model was used on each of the slices separately and then the resulting masks from the slices were concatenated to form a single mask. For all the baseline diffusion models used, the level of added noise was varied and the noise level that gave the best accuracy was chosen to be presented in the table. The 3D Diffusion model showed instability of performance with varying levels of noise (accuracies were 0.132, 0.078, and 0.1 for noise levels 300, 500, and 700, respectively).

\begin{figure}[tbh]
    \centering
\includegraphics[width=0.7\textwidth]{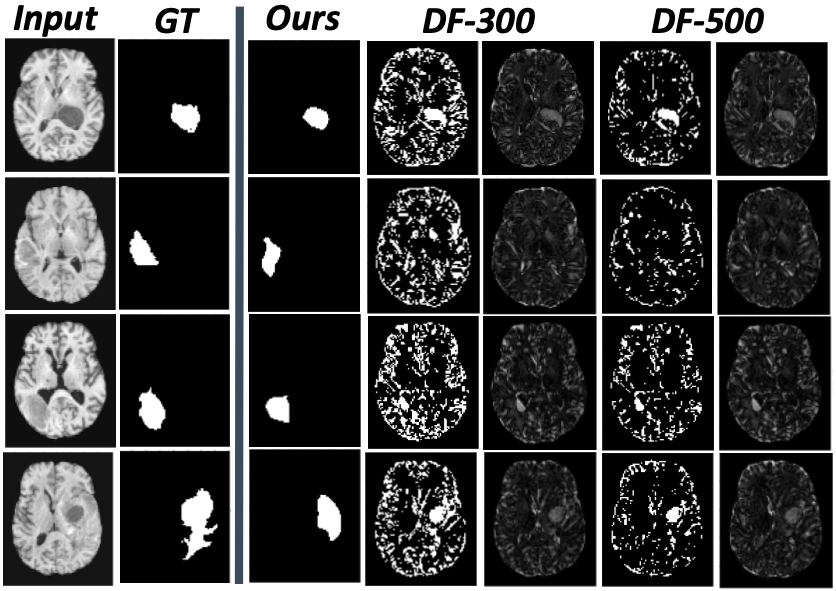}
    \caption{Visual evidence of the effectiveness of our proposed deformation-based lesion segmentation method compared to baseline diffusion models where difference images are binarized based on a chosen threshold.}
    \label{fig:segmentation}
\end{figure}

Table \ref{table:lesion} shows the DICE scores (higher is better) comparison between our proposed deformation based segmentation and diffusion-based segmentation methods. The proposed method showed clear quantitative superiority of performance to the baseline methods without being sensitive to any hyperparameters. Also, qualitative assessment of the segmentation model compared to baseline methods are shown in Figure \ref{fig:segmentation}.

Since anomaly detection is a field that is always active, it is foreseen that other models can emerge that outperform our proposed deformation-based lesion segmentation, and hence can replace this module from our proposed lesion reversal process, which will enhance the overall quality of the process. The proposed method clearly outperformed the existing SOTA 3D methods we included in the comparison. However, we did not compare against all existing anomaly detection methods as this is not the main focus of this work.

\begin{table}[!t]
\centering
\caption{Comparison of Deformation-based (Ours) lesion segmentation against other diffusion-based methods on the BraTS Dataset.}
\label{table:lesion}
\scriptsize 
\setlength\tabcolsep{2pt} 
\renewcommand{\arraystretch}{1.2} 
\begin{tabular}{lcccc}
\toprule
 & \textbf{2D Diffusion} & \textbf{3D Latent Diffusion} & \textbf{3D Diffusion} & \textbf{Ours} \\ 
\midrule
\textbf{Dice Score} & 0.095 & 0.078 & 0.132 & 0.293 \\ 
\bottomrule
\end{tabular}
\end{table}

\subsection{In-painting module Evaluation}

We used two different datasets to evaluate the inpainting property. Our inpainting model was assessed using an in-distribution healthy dataset, IXI, and an out-of-distribution healthy dataset, NFBS. We masked the datasets using random masks extracted from the Brats21 dataset. To compare different methods for the masked Region of Interest (ROI), we calculated the Normalized Mean Squared Error (NMSE) and Structural Similarity Index Measure (SSIM). We compared our fine-tuned model with the RePaint method. Our results, shown in Table \ref{table:inpaint}, indicate that our method slightly improves inpainting performance for both in-distribution and out-of-distribution healthy data in terms of SSIM and Normalized Mean Squared Error (NMSE), although both methods generate a very high SSIM (In-painting examples are provided in the appendix figure \ref{fig:in_results}).

Additionally, we expect the inpainted brain to retain anatomical properties that match those of the original healthy brain. To further evaluate the methods, we generated a synthetic brain lesion mask and applied it to randomly occlude a region of the healthy brain image. The two inpainting methods were then used to restore the occluded region. We compared the anatomical statistics extracted from the restored brain images to those of the original healthy brain image. We processed a healthy brain image using \textit{BrainSuite} to extract key anatomical statistics, including mean cortical thickness (in mm), cerebrospinal fluid (CSF) volume (in mm\(^3\)), gray matter (GM) volume (in mm\(^3\)), white matter (WM) volume (in mm\(^3\)), mid-cortical surface area (in mm\(^2\)), inner cortical area (in mm\(^2\)), and pial cortical area (in mm\(^2\)). We repeat the same process on 10 different brains and masks and report the results averaged over all the brains.

\begin{figure}[tbh]
    \centering
    \includegraphics[width=0.9\textwidth]{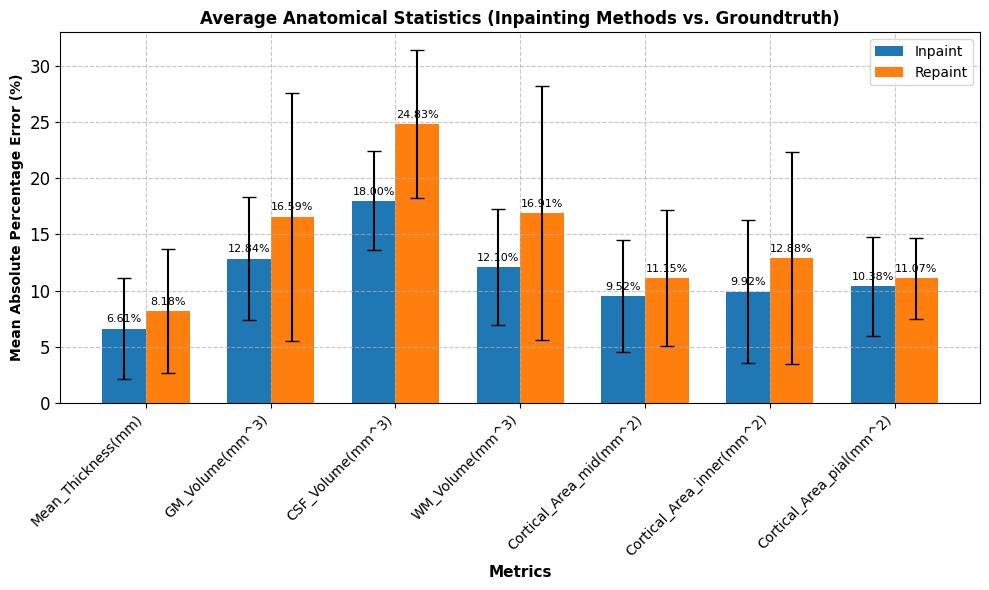}  
    \hfill
    \caption{Comparison of anatomical metrics between inpainting methods and groundtruth, showing mean percentage differences across cortical thickness, volumes, and surface areas.}
    \label{fig:statistics comparison}
\end{figure}

Figure \ref{fig:statistics comparison} presents a comparison of anatomical metrics between the two inpainting methods and the groundtruth. \textit{BrainSuite} extracts these metrics for each region of interest (ROI), and the comparison shown reflects the mean percentage differences averaged across all ROIs. The results indicate that our inpainting method consistently achieves lower percentage differences across key anatomical statistics metrics, including cortical thickness, tissue volumes, and surface areas, compared to the RePaint method. This suggests that our approach more accurately reconstructs brain regions with anatomically consistent content, closely aligning with the structure of healthy brain tissue.

\begin{table}[!t]
\centering
\caption{Comparing in-painting performance of our fine-tuned model with RePaint (SSIM/NMSE).}
\label{table:inpaint}
\scriptsize 
\setlength\tabcolsep{4pt} 
\renewcommand{\arraystretch}{1.2} 
\begin{tabular}{lcccc}
\toprule
\textbf{Dataset/Method} & \multicolumn{2}{c}{\textbf{3D Diffusion-RePaint}} & \multicolumn{2}{c}{\textbf{Ours}} \\ 
\midrule
 & \textbf{SSIM} & \textbf{NMSE} & \textbf{SSIM} & \textbf{NMSE} \\ 
\midrule
\textbf{IXI} & 0.9934 & 0.0034 & 0.9942 & 0.0030 \\ 
\textbf{NFBS} & 0.9918 & 0.0629 & 0.9933 & 0.0429 \\ 
\bottomrule
\end{tabular}
\end{table}

\subsection{Brain Labeling Improvement Evaluation}

\begin{figure}[tbh]
    \centering
\includegraphics[width=\textwidth]{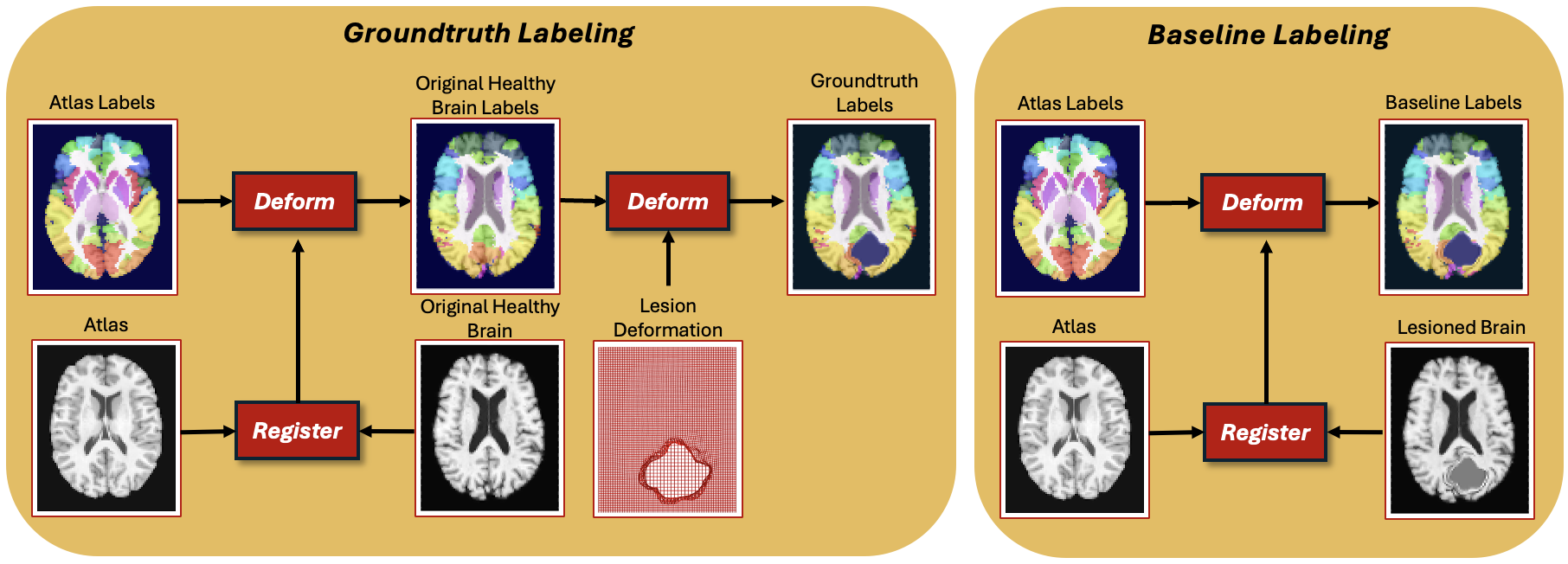}
    \caption{Left: Groundtruth labeling process of lesioned brains. Right: Classic way of labeling lesioned brains}
    \label{fig:Lesion_Labeling}
\end{figure}

The labels of any healthy brain are constructed by learning a mapping between an atlas whose labels are accurately known and the healthy brain. This mapping is then used to map any known spatial information about the atlas to the healthy brain. This process is well-defined and classically used in brain softwares to label and analyze healthy brains. It relies on the fact that there is a point-to-point correspondence between the atlas and the healthy brain, i.e., every point in the atlas can be mapped to a corresponding point in the atlas and vice versa. When labeling a lesioned brain using the same process, this point-to-point correspondence does not exist, as the lesion part of the brain has no correspondence in the atlas. This problem makes the learned mapping between the atlas and the lesioned brain an erroneous one where the error signifies in the lesion neighborhood.

Here, we precisely analyze this problem, and propose a solution that overcomes the broken point-to-point correspondence assumption. To tractably conduct this analysis, and because there is no known groundtruth labels for real lesioned brains, we used our lesion synthesis process to generate brains where the true groundtruth labeling can be found precisely through valid steps. In the left sub-figure in Figure \ref{fig:Lesion_Labeling},  we show the steps in which true groundtruth labels can be found for any lesioned brain; The atlas brain is mapped to the healthy (pre-lesion) brain, and then this mapping is used to transfer the labels of the atlas to the healthy brain. The labels of the healthy brain are then transferred to the lesioned brain through the deformation field that was used to generate the lesion from the lesion core, and then the lesion mask is added on top of the labels. These steps are all valid and do not violate the point-to-point correspondence assumptions. It is worth noting that this process cannot be accurately carried out in real lesioned brain images, because of the difficulty of having the healthy pre-lesion brain image as well as the exact deformation that happened to the brain in the lesion growth process. Hence, the lesion synthesis process is needed. On the right sub-figure in Figure \ref{fig:Lesion_Labeling}, we show the classical process in which lesioned brains are labeled, which relied on an erroneous learned mapping as mentioned earlier.

To this end, we propose a way of improving the classical labeling process by tractably replicating the groundtruth labeling process. Looking at the groundtruth labeling process, we can see that in real scenarios, it is not followed because of the absence of a) the original healthy brain image, and b) the deformation associated with the lesion process. We can see (from Figure \ref{fig:Lesion_Reversal}) that in our proposed lesion reversal process, we estimate both a) the healthy brain image as the final step of the process, and b) the deformation associated with the lesion process (the inverse of the deformation field found in "\(\text{Registration}^{b}\)" in Figure \ref{fig:Lesion_Reversal}).

\begin{figure}[tbh]
    \centering
    \includegraphics[width=0.49\textwidth]{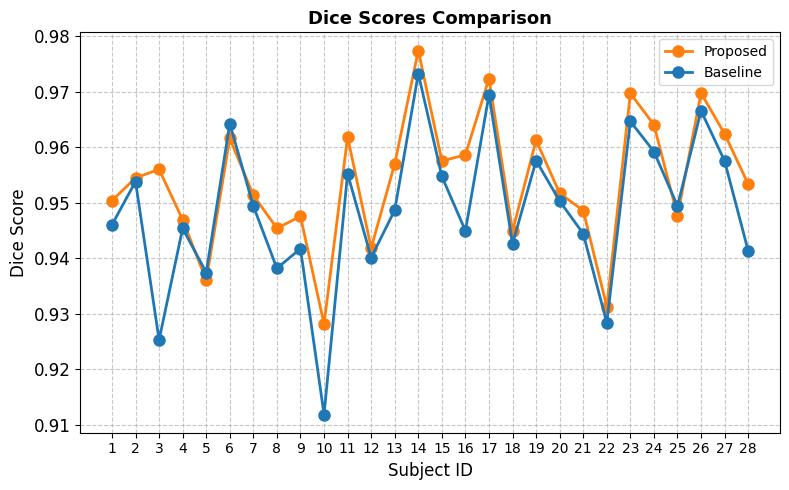}
    \hfill
    \includegraphics[width=0.49\textwidth]{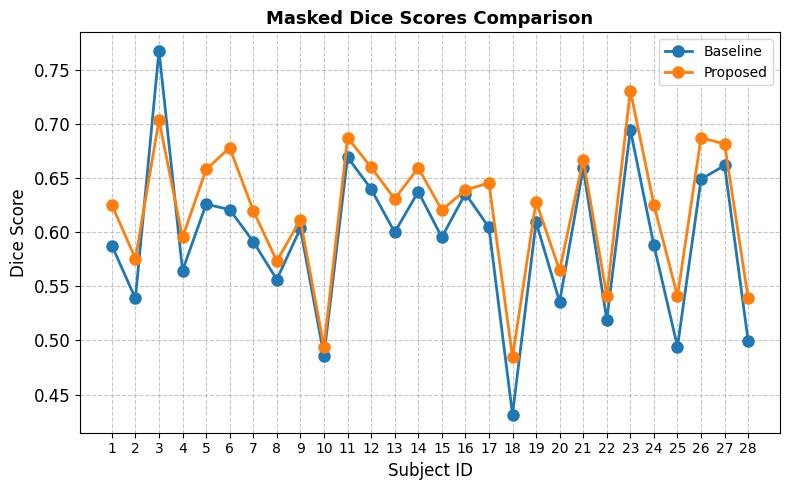}  
    \caption{Left: Dice Score Comparison (On whole Brain). Right: Dice Score Comparison (Around lesion area).}
    \label{fig:comparison}
\end{figure}

\begin{figure}[tbh]
    \centering
    \includegraphics[width=0.5\textwidth]{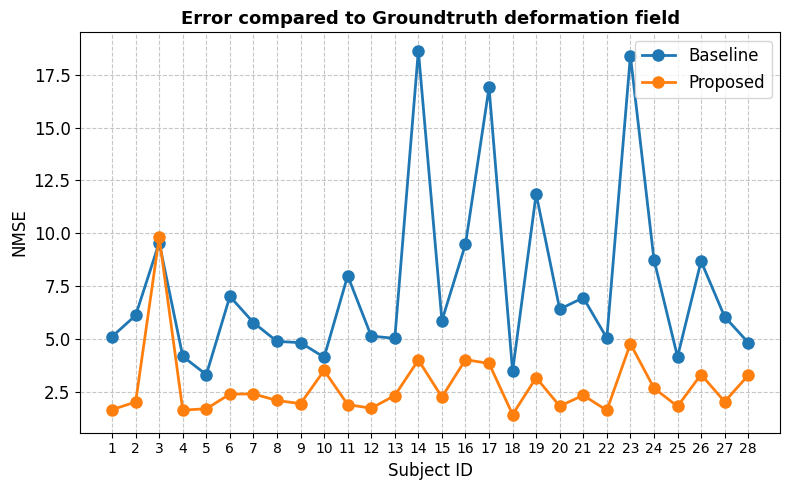}  
    \hfill
    \caption{Comparison between the deformation fields used to generate lesioned brain labels in the proposed labeling process and the baseline process against the groundtruth process.}
    \label{fig:deformation comparison}
\end{figure}

We use our proposed lesion synthesis process to generate 28 lesioned brains, and we replicate the groundtruth labeling process using our estimated healthy brain and lesion deformation. We compare the labeling performance of our proposed method against the baseline labeling process. We compare the DICE scores between our generated labels and the groundtruth labels and the DICE score between the baseline labels and the groundtruth labels as a main comparison metric. Figure \ref{fig:comparison} shows the dice scores comparison between the baseline and proposed processes; on the left, the dice scores are compared in the whole brain, and on the right the dice scores are compared in the areas of the brain that are 10 or less voxels away from the lesion. In addition, we compare the deformation fields used to generate the lesioned brain labels; we calculate the Normalized Mean Square Error (NMSE) between the accumulation of the two deformation fields used in the groundtruth labeling process and the proposed labeling process, against the  NMSE between the the accumulation of the two deformation fields used in the groundtruth labeling process and the deformation field in the baseline labeling process (Figure \ref{fig:deformation comparison}).

Figure \ref{fig:comparison} shows that the proposed labeling method improves the performance of the classical baseline labeling method in all but one subject. This confirms that the proposed method effectively addresses the flawed assumption of point-to-point correspondence in the baseline method, leading to more accurate labeling results.

\subsection{Software Processing of Real Lesioned Brains} 
\label{brain labeling}

We used \textit{BrainSuite} \cite{shattuck2002brainsuite} to process lesioned brain MRIs from Brats21 dataset as well as the estimated healthy brain MRIs that are generated by our proposed lesion synthesis process for each of the lesioned brains.

The anatomical pipeline in \textit{BrainSuite} uses a sequence of processing steps that start with input MRI, skull stripping, bias field correction, tissue classification, inner cortical surface generation, and expansion to generate pial cortical surface representation. This is then followed by surface-constrained volumetric registration to generate surface and volume labels by co-registration to an atlas.

We process and label the lesioned brain using the software, as well as the estimated healthy brain, and then we move the estimated healthy brain and its labels to the lesioned brain space using the inverse of the deformation field found in the "\(\text{Registration}^{b}\)" process in Figure \ref{fig:Lesion_Reversal}. We then compare the labeling of the lesioned brain that was generated by the direct processing versus the labeling that resulted in from our proposed method.

In Figure \ref{fig:br_prc1} (The 3D view is provided in Appendix section \ref{example images} in Figure \ref{fig:br_prc2}), the arrow in the top row shows that the right temporal lobe in the lesioned brain has been mislabelled in the direct labeling method; The medial temporal gyrus and temporal pole have been completely misidentified in the lesioned brain. Additionally, the cortical surface at the temporal lobe has extraction errors and holes due to the lesion. These issues are largely fixed in the bottom row showing the labeling of the cortical surface generated for the lesioned brain by our proposed method.

\begin{figure*}[]
    \centering
\includegraphics[width=0.7\textwidth]{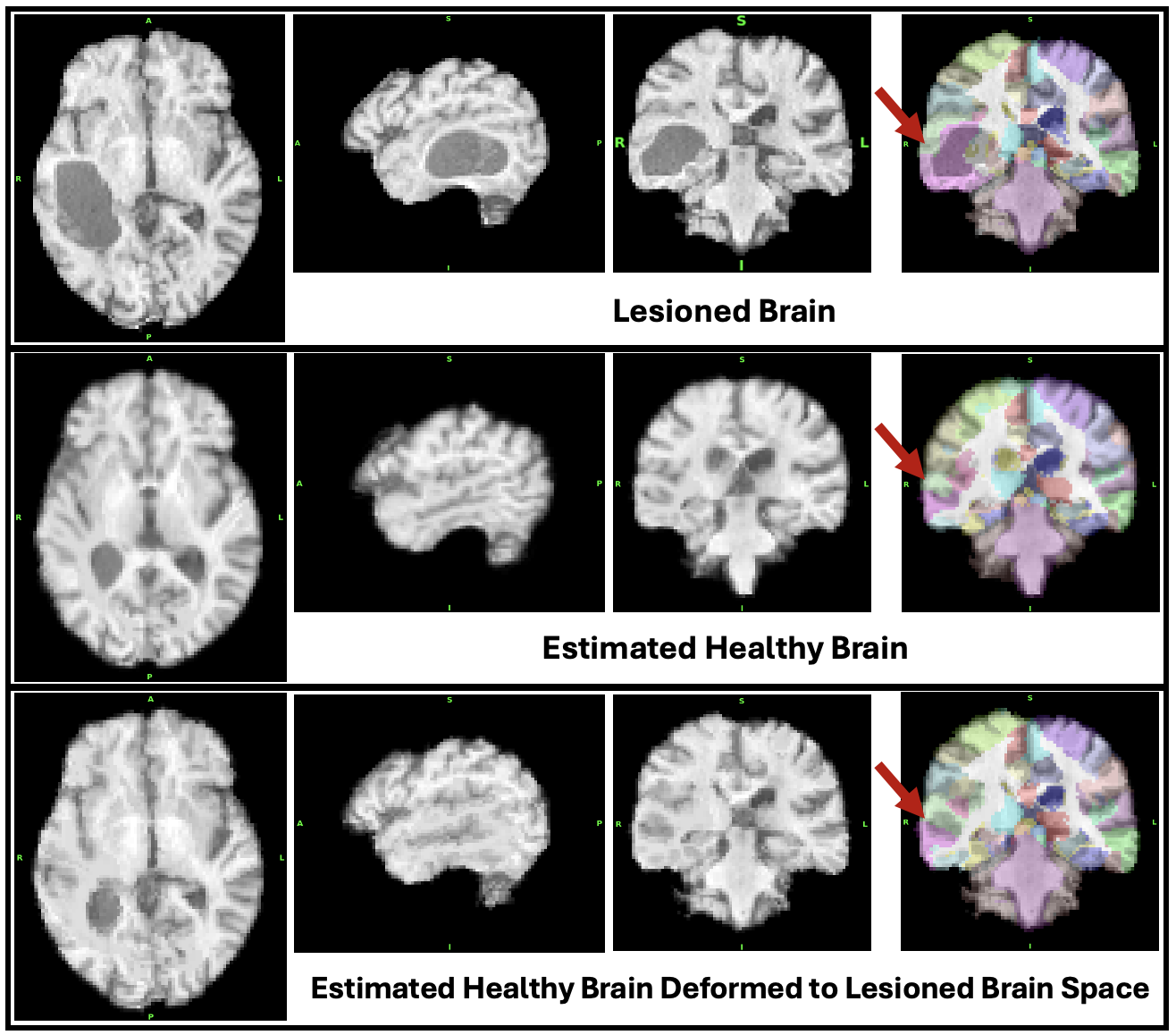}
    \caption{Top row: \textit{BrainSuite} Processing of the lesioned brain. Middle row: \textit{BrainSuite} Processing of the estimated healthy brain. Bottom row: The labels generated for the lesioned brain by moving the labels of the healthy brain to the lesioned brain space. The arrows show that the right temporal lobe in the lesioned brain has been mislabelled in the first row; The medial temporal gyrus and temporal pole have been completely misidentified in the lesioned brain.}
    \label{fig:br_prc1}
\end{figure*}

\section{Conclusion}

In this study, we introduced a novel framework for simulating and reversing brain lesion processes using diffusion models, presenting a comprehensive approach that bridges lesion synthesis, segmentation, and reversal. Our proposed pipeline not only facilitates the generation of realistic synthetic lesions but also enhances the understanding of the effects of lesions on brain tissue, distinguishing between restorable and non-restorable areas. By leveraging a state-of-the-art deformation-based lesion segmentation approach and in-painting techniques, we demonstrated significant improvements in lesion segmentation accuracy and brain labeling, outperforming existing methods that are classically used for these tasks.

The lesion synthesis process provides a potential robust tool for augmenting datasets, allowing better training of machine learning models in medical imaging. The lesion reversal process, on the other hand, offers an innovative solution for reconstructing pre-lesion brain states, aiding in the accurate quantification of lesion impacts on brain regions. This method has practical implications in surgical planning and postoperative monitoring, as it allows clinicians to visualize and assess the original brain structure prior to lesion growth.

Our experiments and evaluations validated the effectiveness of our proposed methodologies and the value of our proposals in both simulated and real-world scenarios, highlighting its potential to improve brain labeling and processing in clinical applications. By accurately estimating the pre-lesion state, our method corrects labeling errors commonly encountered when traditional registration techniques are applied directly to lesioned brains. This advancement underscores the value of integrating lesion simulation and reversal processes into existing clinical workflows, providing a more reliable and informative approach to brain lesion analysis.

In conclusion, our study introduces a framework that simulates and reverses brain lesion processes, improving the understanding and analysis of lesioned brains. Our pipeline offers a practical addition to existing tools, supporting further exploration in brain lesion modeling and clinical applications.


\bibliography{fullpaper}
\newpage

\section{Appendix}
\subsection{Diffusion models}
\label{diff}
In the forward diffusion process, data is gradually corrupted by adding Gaussian noise over a sequence of time steps. This can be represented by the following equation:

\begin{equation}
    \mathbf{x}_t = \sqrt{\alpha_t} \mathbf{x}_{t-1} + \sqrt{1 - \alpha_t} \boldsymbol{\epsilon},
\end{equation}

where:
\begin{itemize}
    \item $\mathbf{x}_t$ is the data at time step $t$.
    \item $\alpha_t$ represents the noise scale at each step.
    \item $\boldsymbol{\epsilon}$ is the random noise vector sampled from a standard Gaussian distribution $\mathcal{N}(0, \mathbf{I})$.
\end{itemize}

\noindent
As $t$ increases, the data $\mathbf{x}_t$ becomes increasingly similar to white Gaussian noise.

The reverse process aims to reconstruct the original data from the noise. It iteratively estimates the noise that was added at each step and removes it. The update rule is:

\begin{equation}
    \mathbf{x}_{t-1} = \frac{1}{\sqrt{\alpha_t}} \left( \mathbf{x}_t - \frac{1-\alpha_t}{\sqrt{1-\bar{\alpha}_t}} \boldsymbol{\epsilon}_\theta(\mathbf{x}_t, t) \right),
\end{equation}

where $\bar{\alpha}_t$ is the product of all previous $\alpha$ values up to time $t$, and $\boldsymbol{\epsilon}_\theta(\mathbf{x}_t, t)$ is the predicted noise by the neural network with parameters $\theta$.

\subsection*{Optimization Objective}

The model is trained to minimize the difference between the true noise and the predicted noise using a loss function :

\begin{equation}
\label{eq:opt-obj}
    \mathcal{L}(\theta) = \mathbb{E}_{t, \mathbf{x}_0, \boldsymbol{\epsilon}} \left[ \| \boldsymbol{\epsilon} - \boldsymbol{\epsilon}_\theta(\mathbf{x}_t, t) \|^2 \right],
\end{equation}

where $\mathcal{L}(\theta)$ is the loss function for the neural network parameters $\theta$, and the expectation is taken over various $t$, the original data $\mathbf{x}_0$, and the noise $\boldsymbol{\epsilon}$.

\subsection{Preprocessing}
\label{preprocessing}
To align all MRI scans, we register the brain scans to the SRI24-Atlas (\cite{rohlfing2010sri24}) by affine transformations. Next, we apply skull stripping with HD-BET \cite{isensee2019automated}. Note that these steps are already applied to the BraTS21 data set by default. Subsequently,
we remove black borders, leading to a fixed image size of [192 × 192 × 160] voxels. Lastly, we perform a bias field correction. To save computational resources, we reduce the volume resolution by a factor of two, resulting in [96×96×80] sized image. To standardize intensity we divided each image by the value of the largest peak in its histogram.
\subsection{Datasets}
\label{dataset}
\textbf{UK Biobank (UKB):}
The UKB is a large-scale biomedical database and research resource containing in-depth genetic and health information from half a million UK participants. The study aims to improve the prevention, diagnosis, and treatment of a wide range of serious and life-threatening illnesses. We utilized a subset of 5,000 random participants with T1w images out of 45,564 subjects, all of whom were healthy individuals aged between 44 and 82 years.

\noindent
\textbf{Human Connectome Project (HCP):}
The HCP aims to construct a detailed map of the human brain's structural and functional connections. Our study incorporated 1,113 subjects from the HCP dataset, aged 22-35 years, all of whom underwent minimal preprocessing. This preprocessing includes spatial artifact/distortion removal, surface generation, cross-modal registration, and alignment to standard space, ensuring a high standard of data quality for our analyses.

\noindent
\textbf{Cambridge Centre for Ageing and Neuroscience (Cam-CAN):}
The Cam-CAN dataset is a resource for studying cognitive and brain aging over the adult lifespan. We included T1-weighted weighted MRI images from 653 adults aged 18-88. This subset, known as the "CC700", provides a wide age range.

\noindent
\textbf{IXI Dataset:}
The IXI dataset consists of 560 pairs of T1 and T2-weighted brain MRI scans acquired in three different London hospitals. From this dataset, we used 158 samples for testing and partitioned the remaining into 358 training samples and 44 validation samples, following Behrendt et al., 2023 /cite{behrendt2023patched}. This dataset enhances the diversity of our data, adding more depth to our analysis across different populations and imaging settings.

\noindent
\textbf{Neurofeedback Skull-stripped (NFBS) Dataset}\footnote{http://preprocessed-connectomes-project.org} Available from the Preprocessed Connectomes Project repository, includes 125 manually skull-stripped T1-weighted anatomical MRI scans of individuals aged 21 to 45 years. Each scan in this dataset has been meticulously checked to ensure no brain abnormalities, as confirmed by a board-certified neuroradiologist. 

\noindent
\textbf{BRaTS21 Dataset} The
Tumor Segmentation Challenge 2021 (BraTS21) dataset (Baid et al., 2021) comprises 1251 brain
MRI scans of four different weightings, including T1, T1-CE, T2, and FLAIR, with 100
random samples were used in this study. Each scan in this dataset is accompanied by an expert
annotations in the form of pixel-wise segmentation maps, providing a detailed ground truth
for evaluating the model’s performance.

Each dataset contributes a unique demographic and technical perspective to our study, providing a robust and comprehensive dataset for training our models. The integration of these diverse datasets is expected to enhance the generalizability and reliability of our findings.

To assess the performance and adaptability of our inpainting modules, we employed Two distinct datasets, each reflecting a unique population type: the test subset of the IXI dataset, and the Neurofeedback Skull-stripped (NFBS) dataset.
The IXI dataset served as a benchmark for evaluating the inpainting model's performance on data similar to the training distribution, providing a baseline for expected performance under ideal conditions. The NFBS dataset was selected to test the model's performance on out-of-distribution healthy brain images. This provides insight into the model's generalizability and its ability to maintain performance when confronted with healthy but structurally different brain images from those seen during training.

\subsection{Lesion Synthesis Algorithm}
\label{synthesis algorithm}

\floatname{algorithm}{Algorithm}
\renewcommand{\algorithmicrequire}{\textbf{Input:}}
\renewcommand{\algorithmicensure}{\textbf{Output:}}
\newcommand{\algorithmicinit}{\textbf{Initialize:}}
\newcommand{\INIT}{\item[\algorithmicinit]}

\begin{algorithm}
\caption{Synthesize Lesioned Brain Images}
\label{alg:lesion_synthesis}
\begin{algorithmic}[1]
    \Require Healthy brain image $I_{healthy}$, Lesion location $p_{lesion}$, Final lesion size $s_{final}$, Deformation severity $\alpha$
    \Ensure Synthesized brain image with lesion $I_{lesion}$

    \State \textbf{Step 1: Core Lesion Placement}
    \State Generate a small spherical lesion core $L_{core}$ at the specified location $p_{lesion}$ within the white matter region. 
    \State Initialize $L_{core}$ with intensity values averaged from surrounding brain tissue to ensure realistic integration.

    \State \textbf{Step 2: Lesion Core Mask}
    \State Apply Gaussian smoothing to soften the edges of $L_{core}$. Perturb the core's surface with Perlin noise to create a more irregular, blob-like structure. This simulates natural lesion variability. 

    \State \textbf{Step 4: Final Lesion Mask Placement}
    \State Generate a larger blob $L_{final}$ that represents the final lesion size $s_{final}$. This is positioned to overlap with the center of the lesion core, simulating the lesion's growth.

    \State \textbf{Step 5: Displacement Field Calculation}
    \State Find a dense displacement field (DDF) that deforms $I_{lesion}$ such that the core lesion mask $L_{core}$ expands to match the final lesion mask $L_{final}$. The displacement is controlled by the deformation severity parameter $\alpha$.
    \State Ensure the DDF is diffeomorphic by integrating the dense velocity field (DVF), maintaining smooth and realistic tissue deformation.

    \State \textbf{Step 6: Final Image}
    \State Apply the DDF to $I_{lesion}$, modeling tissue displacement and generating the final brain image $I_{final}$, where the lesion has grown to the desired size and shape.
    
    \State \Return $I_{final}$
\end{algorithmic}
\end{algorithm}

 \subsection{In-painting and Atlas Mapping Examples}
 \label{example images}

\begin{figure*}[tbh]
    \centering
\includegraphics[width=1\textwidth]{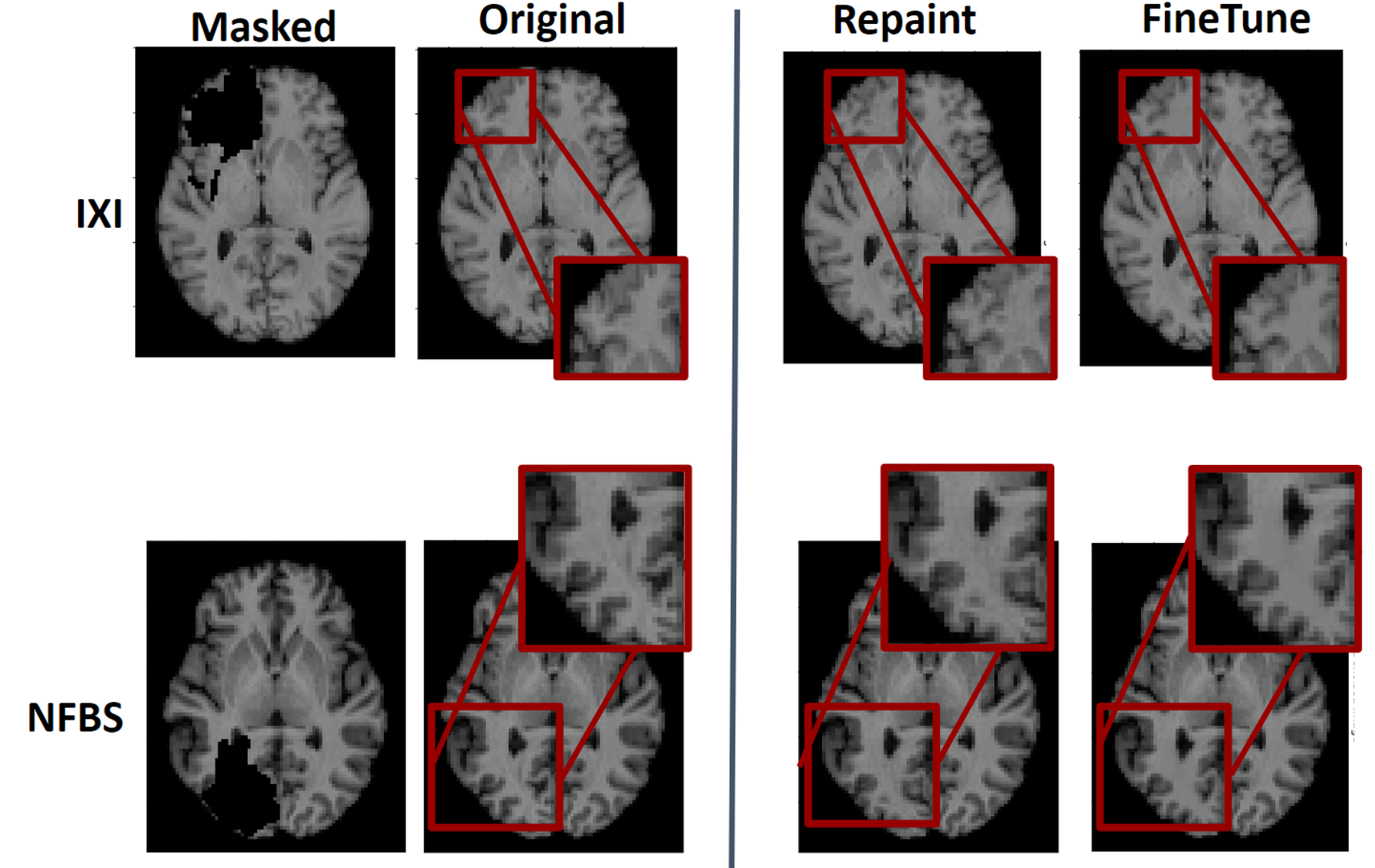}
    \caption{ inpainting results comparing RePaint and our fine-tuned model}
    \label{fig:in_results}
\end{figure*}

\begin{figure*}[]
    \centering
\includegraphics[width=1\textwidth]{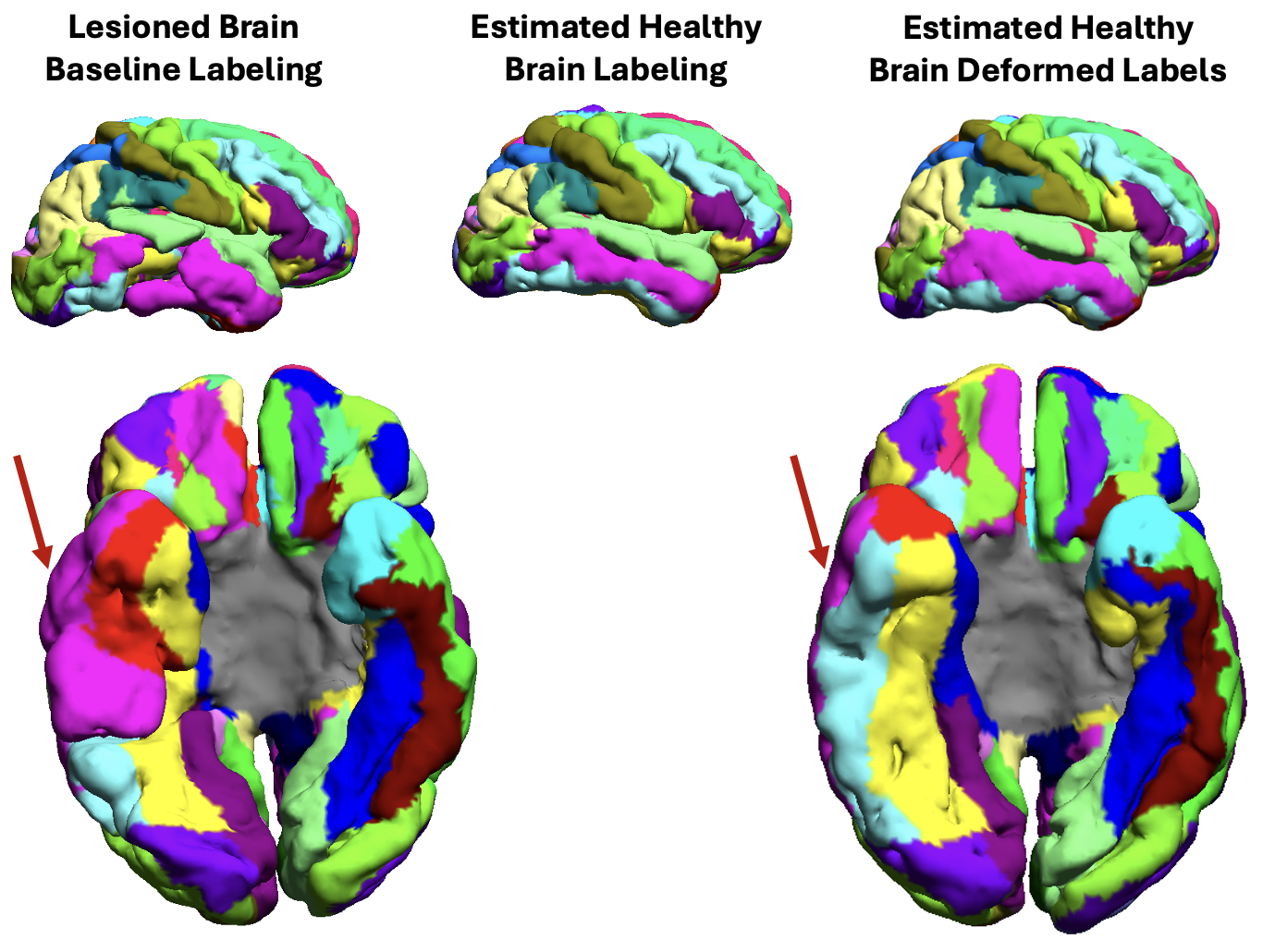}
    \caption{\textit{BrainSuite} Processing of the lesioned brain, Estimated healthy brain, and estimated healthy brain moved back to lesioned brain space in 3D}
    \label{fig:br_prc2}
\end{figure*}

\end{document}